\documentclass[A4paper, 11 pt, conference]{ieeeconf}  

\IEEEoverridecommandlockouts       
\usepackage{geometry}
\usepackage{graphics} % for pdf, bitmapped graphics files
\usepackage{epsfig} % for postscript graphics files
\usepackage{mathptmx} % assumes new font selection scheme installed
\usepackage{times} % assumes new font selection scheme installed
\usepackage{amsmath} % assumes amsmath package installed
\usepackage{amssymb}  % assumes amsmath package installed
\usepackage{subfigure}
\usepackage{algorithm} %format of the algorithm
\usepackage{algorithmic} %format of the algorithm
\usepackage{multirow} %multirow for format of table
\usepackage{tabu}
\usepackage{makecell}
\usepackage{threeparttable}

\begin{document}

% Change to your title
\title{\LARGE \bf
Robust Edge-Direct Visual Odometry based on CNN edge detection and Shi-Tomasi corner optimization}

\author{Kengdong Lu$^{1, 3,*}$, Jintao Cheng$^{1, *}$, Yubin Zhou$^1$, Juncan Deng$^1$, Rui Fan$^2$, Kaiqing Luo$^{1, 3}$} % <-this % stops a space
% \thanks{
% 1. Author is with the School of Physics and Telecommunication Engineering, South China Normal University, Guangzhou 510006, P. R. China.
% 2. Author is with the College of Electronic and Information Engineering, Tongji University, Shanghai 201804, P. R. China.
% 3. Author is with the Shanghai Research Institute for Intelligent Autonomous Systems, Shanghai 201210, P. R. China.
% 4. Author is with the Guangdong Provincial Engineering Research Center for Optoelectronic Instrument, South China Normal University, Guangzhou 510006, P. R. China.
% kdlu@m.scnu.edu.cn, chengjintao@unity-drive.com, zhouyubin@m.scnu.edu.cn, rui.fan@ieee.org, kqluo@scnu.edu.cn%%
% }}
\maketitle
\let\thefootnote\relax\footnotetext{ Corresponding:Kaiqing Luo. (email:kqluo@scnu.edu.cn) }
\let\thefootnote\relax\footnotetext{ This work was supported in part by the National Science Foundation of China (NSFC) - Guangdong big data Science Center Project under Grant U1911401,National Natural Science Foundation of China (61975058); Natural Science Foundation of Guangdong Province (2019A1515011401); the Science and Technology Program of Guangzhou (2019050001); and the Science and Technological Plan of Guangdong Province, China (2019B090905005), and in part by the South China Normal University National Undergraduate Innovation and Entrepreneurship Training Program under Grant 202010574050 and 202110574048.  }
\let\thefootnote\relax\footnotetext{ 1. K. Lu, J. Cheng, Y. Zhou, J. Deng, K. Luo are with the School of Physics and Telecommunication Engineering, South China Normal University, Guangzhou 510006, P. R. China. }
\let\thefootnote\relax\footnotetext{ 2. R. Fan is with the College of Electronic and Information Engineering, Tongji University, Shanghai 201804, P. R. China. as well as Shanghai Research Institute for Intelligent Autonomous Systems, Shanghai 201210, P. R. China. (email:rui.fan@ieee.org). }
\let\thefootnote\relax\footnotetext{ 3. K. Lu, K. Luo are with the Guangdong Provincial Engineering Research Center for Optoelectronic Instrument, South China Normal University, Guangzhou 510006, P. R. China. }
\let\thefootnote\relax\footnotetext{ *The authors contributed equally to this work. }

\maketitle 
\thispagestyle{empty}

\begin{abstract}
In this paper, we propose a robust edge-direct visual odometry (VO) based on CNN edge detection and Shi-Tomasi corner optimization. Four layers of pyramids were extracted from the image in the proposed method to reduce the motion error between frames. This solution used CNN edge detection and Shi-Tomasi corner optimization to extract information from the image. Then, the pose estimation is performed using the Levenberg-Marquardt (LM) algorithm and updating the keyframes. Our method was compared with the dense direct method, the improved direct method of Canny edge detection, and ORB-SLAM2 system on the RGB-D TUM benchmark. The experimental results indicate that our method achieves better robustness and accuracy.
\end{abstract}

\section{Introduction}

The visual odometry \cite{ref1} (VO), which can estimate the camera motion between the adjacent images and the planning of local maps, and plays a vital role in the synchronous positioning and map construction technology \cite{ref2,ref34,ref3}. It has many practical applications in robotics, such as autonomous driving, navigation, augmented reality, and three-dimensional reconstruction.

In the past few decades, there has been a lot of work on VO \cite{ref35, ref36}. The idea of VO was first proposed by Moravec et al. \cite{ref4} Follow-up related work can be divided into feature methods and direct methods.  Feature-based (indirect) methods are to extract some salient features from dense image data for calculation. The VO system using the feature method runs stably with a low computational cost, and it is robust to factors such as illumination, image noise, etc. The famous ORB-SLAM\cite{ref5} and ORB-SLAM2\cite{ref6} are based on the feature point method. However, the VO system using the feature method is not suitable for scenes that lack features\cite{ref32}, such as gradual images. Compared with the indirect method, the direct method uses the gray information of all pixels in the image or a specific sub-region to calculate the camera's motion. The VO system using the direct method uses pixel gradients and does not need to use feature points in the image\cite{ref33}. It makes full use of image information, which is conducive to the realization of visual applications such as building dense maps\cite{ref7}. However, the direct method has a large amount of calculation and is unsuitable for large motions. Meanwhile, the direct method requires that the image must meet the assumption that the gray pixel value is constant, and this assumption will be destroyed due to illumination and other reasons. According to the number of pixels used, the direct method can be divided into three types: sparse, semi-dense\cite{ref8}, and dense.

\begin{figure}
	\begin{center}
		\centering
		\includegraphics[width=0.4\textwidth]{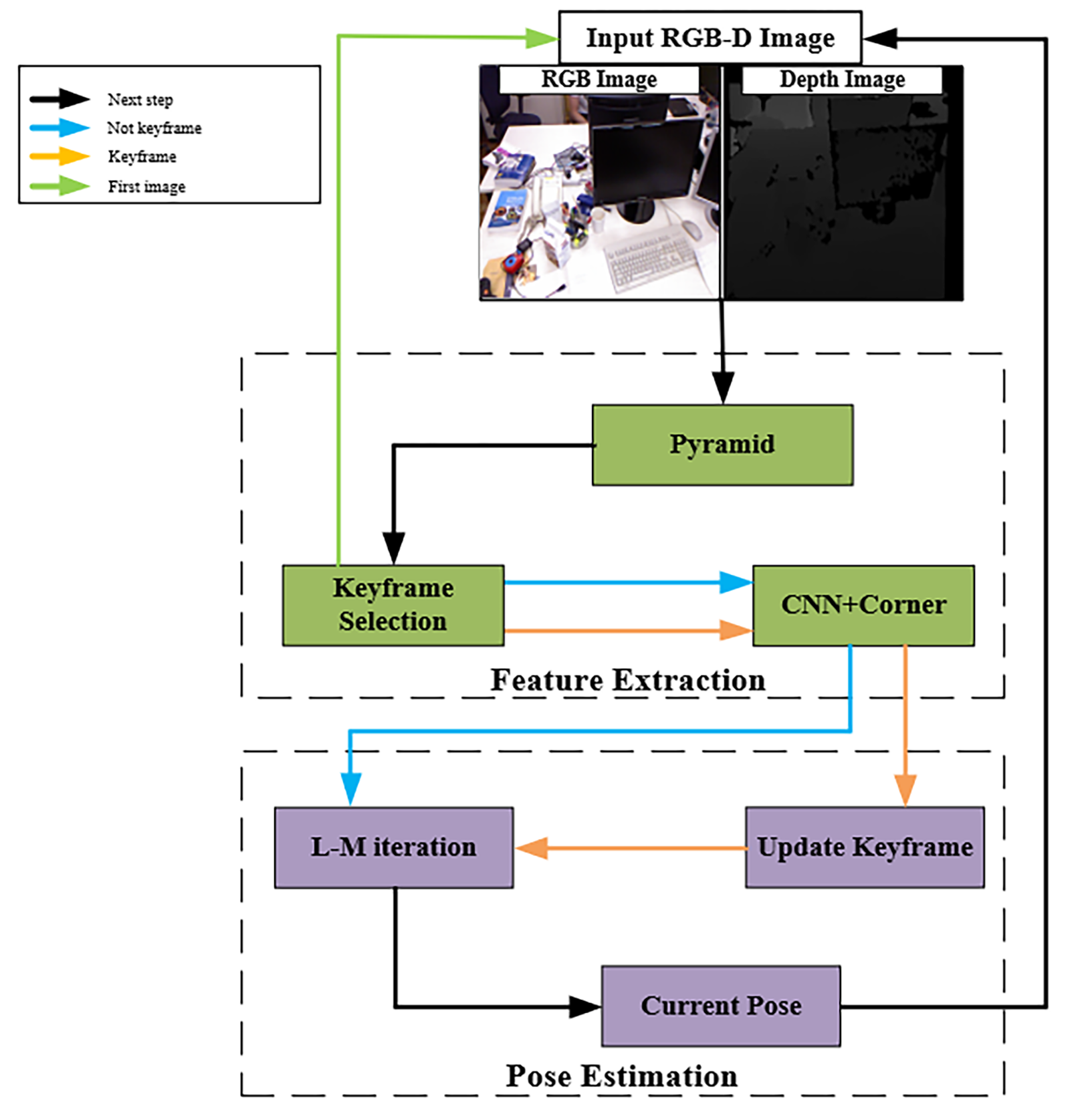}
		\caption{ Flowchart of the our method. Pyramid denote extracts four-layer pyramid, CNN+Corner denote CNN edge detection and  Shi-Tomasi corner optimization. L-M iteration denote Levenberg-Marquardt to calculate the pose of two frames iteratively. }
		\label{fig1}
	\end{center}
\end{figure}

In this paper, we propose a robust edge-direct visual odometry based on CNN edge detection and Shi-Tomasi corner optimization. In the evaluation results of the standard TUM RGB-D benchmark data set, we compared this method with the ORB-SLAM2 system. RPE and ATE data results show that this method has good accuracy in multiple data set sequences.

The contributions of the work:
\begin{itemize}
   \item Our method uses advanced CNN edge detection algorithms. Combining CNN edge detection and Direct-VO into Edge-Direct VO achieves excellent performance compared to other algorithms in evaluating the TUM RGB-D benchmark.
   \item  In our Edge-Direct VO, we propose an improved Shi-Tomasi angle optimization for edge maps, which optimizes the pose estimation of VO to make the whole method more accurate and robust.
   \item Keyframes can effectively reduce accumulated errors. Our method uses a dual mechanism combining periodicity and motion amplitude to update keyframes, which significantly improves the accuracy of experimental results.
\end{itemize}

\section{Related Work}  % TODO check 大小写
In this section, we will briefly introduce edge detection and outline some of the vital work in the field of VO.

\subsection{Edge Detection}
Edge detection is a focused area in computers, and it is a challenging subject. The traditional Sobel operator and Canny operator edge detectors have a wide range of applications \cite{ref37, ref38, ref39}. However, these detectors only take into account sharp local changes and look for edges from these features, especially sharp changes in color, brightness etc. In 2013, people began to use machine learning-based methods to learn how to combine the features of color, brightness, and gradient for edge detection, such as gPb\cite{ref9} and StructuredEdge\cite{ref10}. The Berkeley research group has established an internationally recognized evaluation set called Berkeley Segmentation Benchmark to evaluate the edge detection algorithm better. The BSDS500 data set\cite{ref11} consists of 500 natural images with hand-labeled edges, which are used to train learning-based edge detection techniques\cite{ref9,ref10}. With the rapid development of deep learning, neural network-based edges detection has become important. This method essentially regards the edge detection process as an edge pattern recognition process. Due to its excellent performance in various edge test data sets, more and more edge algorithms based on deep learning have been considered. In this article, we choose a new CNN edge detection algorithm DexiNed\cite{ref12}, which consists of a bunch of learning filters, receives the image as input, and then predicts the edge map with the same resolution. It shows a better performance.

\begin{figure}
   \begin{center}
      \centering
      \includegraphics[width=0.45\textwidth]{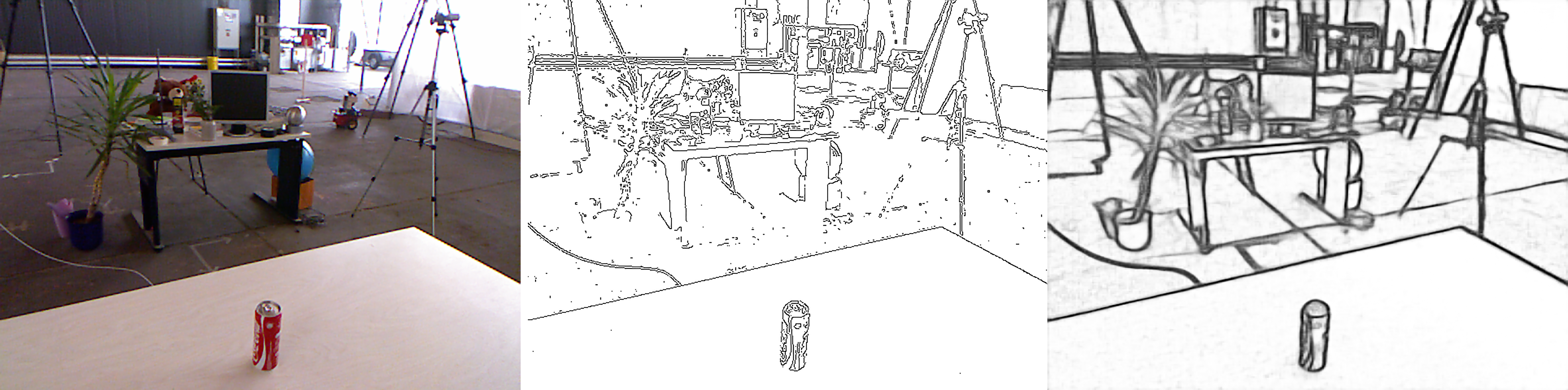}
      \caption{ Original picture (left), Canny edge detection(middle),CNN edge detection (right). }
      \label{fig2}
   \end{center}
\end{figure}

\subsection{Visual Odometry(VO)}
\begin{enumerate}
   \item Feature-based VO: The traditional feature-based (indirect) method extracts features (such as SIFT, SURF, ORB), finds the correspondence between images, and then tracks them in the sequence to estimate the camera's movement. Due to the high inaccuracy in feature extraction and matching, this algorithm must calculate the basic or homography matrix in the RANSAC loop. In terms of using indirect methods, better-known systems are ORB-SLAM, ORB-SLAM2, and Parallel Tracking and Mapping (PTAM)\cite{ref13}. In addition, RGBD-SLAM\cite{ref14} is also a feature-based (indirect) mapping system.
   \item Direct VO: The direct method can optimize the geometry of the image intensity without the need for feature extraction. Therefore, it can work in some environments without texture. Direct method has been widely used in different sensors, such as DVO\cite{ref15} of RGB-D camera and LSD SLAM\cite{ref16} of monocular camera. The core idea is to maintain the semi-dense mapping of keyframes and then minimize the luminosity error. This is a highly non-convex function, so a good initialization is required. However, the direct (feature-free) method estimates camera motion from image data, thus ignoring feature extraction and robust correspondence matching. That‘s the reason why they are usually restricted to small inter-frame motion, which can only be circumvented to a certain extent by image pyramids. Most of these methods rely on the optical consistency assumption\cite{ref17}, which makes them particularly vulnerable to changes in lighting conditions.
   \item Edge based VO: Edge is another vital feature besides points. Compared with points, edges can retain more information and are more robust to changes in light. The edge based methods are indirect method and direct method, but the camera motion estimation does not need correspondence, but the camera motion estimation does not need to correspond. The general idea is to minimize the distance between the edge of one frame and the reprojected edge of another frame.
\end{enumerate}

Eade and Drummond\cite{ref4}, Klein, and Murray\cite{ref13} have mentioned edge-based SLAM. They gathered edge pixels into edges, parameterized and processed them as features. However, there was a problem that the edges are difficult to match. Tarrio and Pedre\cite{ref18} proposed an edge-based monocular camera VO, and they searched along the normal direction to match the edge. It also speeds up the matching step by pre-calculating the distance transform (DT) in a frame\cite{ref19} to speed up this matching step. This idea was adapted by Kuse and Shen\cite{ref20} applied in their RGB-D Direct Edge Alignment (D-EA), They observed and optimized the camera motion based on gradient estimation. Fabian Schenk and Friedrich Fraundorfer\cite{ref21} proposed REVO, a real-time robust RGB-D VO method based on the edge. They use the distance transformation on the edge to reduce the Euclidean geometric error, and use the edges of Se and hed to move relative to the camera. Yang and Scherer\cite{ref22} proposed direct odometry based on points and lines, where for a textured environment, the estimated camera pose is comparable to ORB SLAM\cite{ref17}. Yi and Laurent's Canny-VO’s\cite{ref23} RGB-D visual odometer proposed two alternative methods of distance transformation commonly used in edge registration: Approximate Nearest Neighbour Fields and Oriented Nearest Neighbour Fields, enhanced the efficiency and accuracy of 3D-2D edge alignment. Mingsu\cite{ref24} proposed a monocular vision method based on edge feature detection and deep recurrent convolutional neural networks, embedding traditional geometric algorithms into virtual reality based on deep learning to enhance the influence of image edge feature information.

\subsection{Corner Detection}
The corner point is a critical geometric element in space, which can retain the image characteristics. There will be corner points around the high rate of change of its gradient value. The current mainstream corner detection algorithm is a corner detection algorithm based on grayscale changes. Moravec et al\cite{ref5}. proposed to calculate the grayscale difference and select a minor grayscale variance as the corner response value for non-maximum value suppression to determine the angle. Point. Harris et al\cite{ref25}. based on the Moravec algorithm proposed to extract the corner points through the differential operation and autocorrelation function. Shi and Tomasi et al\cite{ref26}. improved the corner response function of the Harris operator, proposed to extract The Shi-Tomasi\cite{ref26} operator with a more uniform and reasonable feature point distribution can significantly improve the corner extraction effect. Rosten et al\cite{ref27}. proposed the classic FAST and FAST-ER image corner detection methods. It assumes the corner points and performs threshold calculation to verify whether it is an actual corner point.

\section{Method}
In this section, we briefly outline the proposed method. The flow chart of our entire method is shown in Figure \ref{fig1}.

% TODO 大小写问题
% TODO 怎么下面这段用Fig 2，上面的用了Figure 1，是不是要统一一下
\subsection{CNN edge detection}
On the VO based on the sparse method\cite{ref28}, a framework needs to learn geometric feature representation to solve the pose estimation. Compared with the classical edge detection algorithm such as Canny and the machine learning algorithm SE\cite{ref14, ref15}, the extracted images have the problems of too much noise so that it cannot obtain a high accuracy pose estimation. Therefore, an edge detection method based on CNN edge detection\cite{ref12} is utilized in this paper, which is shown in Fig \ref{fig2}. The CNN edge detection has the advantages of extracting high precision features to solve the process of motion.

The CNN edge detection\cite{ref12} takes the original monocular image as input then predicts an edge-map with the exact resolution. It can be seen as two sub-networks, including the dense extreme inception network and the up-sampling block.

In this work, the CNN edge detection\cite{ref12} is used by extracting edge features in RGB-D images. We subtract the mean RGB values of the original RGB-D images and study various edge detection methods and depicted in Figure \ref{fig2}. It can be seen that many edges with a significant distance(outliers) are extracted using the method. Thus, we decide to choose the CNN edge detection. 

\begin{figure}
   \begin{center}
      \centering
      \includegraphics[width=0.45\textwidth]{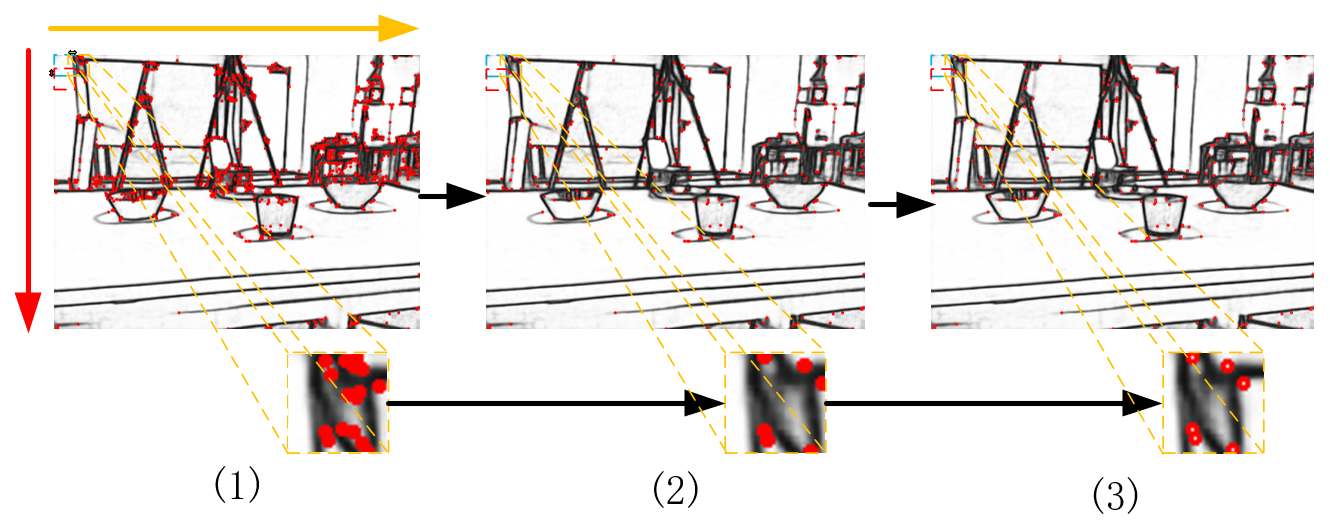}
      \caption{ The process of Shi Tomasi corner optimization. }
      \label{fig3}
   \end{center}
\end{figure}

% TODO 大小写问题
\subsection{Shi-Tomasi Corner Optimization}
The corner extraction of image is easily affected by the image environment, so we use the edge image as a method. A large number of corners will be generated where the gray value changes drastically. This will significantly increase the number of post-processing calculations and affect the accuracy of the subsequent LM iterative calculation of pose. Then Shi-Tomasi corner optimization is performed on areas with too dense corner points to improve the quality of corner points and the speed of computation. The specific steps are as follows:
\begin{enumerate}
   \item Perform Shi-Tomasi\cite{ref26} corner detection on the RGB edge images in the TUM dataset. A fixed-size window $ W(x,y) $ is set, and its pixel gray value is $ I(x, y) $. The window is moved to the x and y directions by a small displacement u, v, and the pixel corresponding to the new position is gray the degree value is $ I(x + u, y + v) $. The change value of the gray value of this movement can be obtained $ [I(x + u, y + v) - I(x, y)] $. Let Gaussian kernel The window functions whose function $ \omega(x, y) $ is $ W(x, y) $ represents the weight of each pixel in the window.	The resulting grayscale value change $ E(u,v) $ can be expressed as:
   \begin{equation}
      \begin{aligned}
      E(u, v) & \approx \sum_{(x, y)} \omega(x, y)\left[I_{x} u+I_{y} v\right]^{2} \\
      &=\sum_{(x, y)} \omega(x, y)[u, v]\left[\begin{array}{cc}
      I_{x}^{2} & I_{x} I_{y} \\
      I_{x} I_{y} & I_{y}^{2}
      \end{array}\right]\left[\begin{array}{c}
      u \\
      v
      \end{array}\right] \\
      &=[u, v] M\left[\begin{array}{c}
      u \\
      v
      \end{array}\right]
      \end{aligned}
      \label{eq1}
   \end{equation}

   \begin{equation}
      M=\sum_{(x, y)} \omega(x, y)\left[\begin{array}{cc}
      I_{x}^{2} & I_{x} I_{y} \\
      I_{x} I_{y} & I_{y}^{2}
      \end{array}\right]
      \label{eq2}
   \end{equation}

   \item It can be seen from the Eq (\ref{eq2}). that the magnitude of the gray value change depends on the matrix M. To find the window that causes a large gray value change, the eigenvalues $ \lambda_1 $ and $ \lambda_2 $ of the $ M $ matrix can be used to calculate the corner response function $ R $ corresponding to each window. Setting a threshold $ \tau_c $, and the corner will meet the following conditions:

   \begin{equation}
      R=\min \left(\lambda_{1}, \lambda_{2}\right)>\tau_{c}
      \label{eq3}
   \end{equation}
   Set a 20x20 pixel window, slide the image from left to right and from top to bottom, and set the sliding step to 20. Count the number of corner points in the sliding window. If the corner point is greater than the preset threshold, the corner point in the window is cleared, and the preset Shi-Tomasi corner point extraction scheme is set.

   \item Draw a circle at the optimized corner coordinates and assign pixel points at the center of the circle to form a small concentric circle area. The process is shown in Figure \ref{fig3}.
\end{enumerate}

Through our practice, Through our practice, when we use LM algorithm to iteratively optimize the front and rear frames and collect the Shi Tomasi corner improvement scheme proposed earlier, the accuracy of RPE and ate has been significantly improved.

\subsection{Keyframe Selection}
\begin{figure}
   \begin{center}
      \centering
      \includegraphics[width=0.47\textwidth]{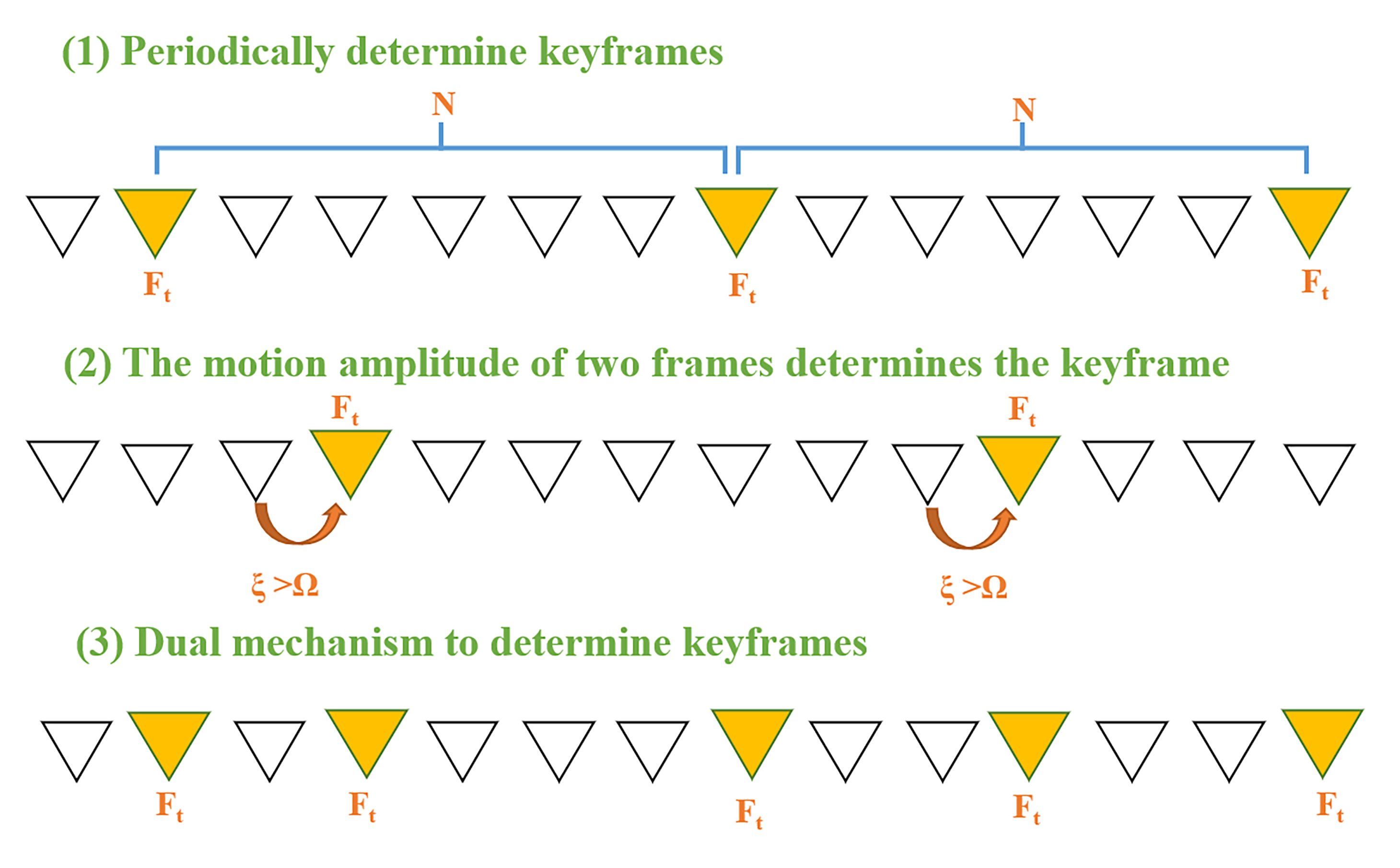}
      \caption{ (1) Our method determines the keyframe through a fixed period. (2) Our method determines the keyframe through the motion amplitude. (3) The final keyframe is determined by a dual mechanism. $ F_t $ represents the key frame, $ {\xi} $ represents the camera movement, ${\Omega}$ and represents the preset movement amplitude. }
      \label{fig4}
   \end{center}
\end{figure}

Our proposed VO method introduced keyframes as reference frames for some new frames, thereby reducing accumulated errors. The selection of keyframes generally depends on the type of VO algorithm. Based on feature points\cite{ref17} usually restrict that a significant number of frames pass, on the order of tens of frames. The method we adopt is to update the keyframe by combining the dual mechanism of periodicity\cite{ref27} and motion amplitude\cite{ref30}. Among them, the updated keyframe is shown in Figure \ref{fig4}.

\subsection{Pose estimation based Levenberg-Marquardt}
To solve the pose estimation, we employ a coarse-to-fine approach Levenberg-Marquardt (LM)\cite{ref29} minimization to avoid the pose optimization falling into a locally optimal solution. Meanwhile, the selection of the image pyramid plays a vital role in the system performance liked \cite{ref27} and \cite{ref30}. Thus, we choose a three levels image pyramid because it performs well in most data sets.

In our method, the edges $ E_t $ will be detected in each original frame $ F_t $ from the intensity It as:
\begin{equation}
   E_{t}=E\left(I_{t}\right)
   \label{eq4}
\end{equation}

We estimate the relative pose $ P_{KC} $ from a current frame $ Fc $ to keyframe $ Ft $ by minimizing the sum over all edge distance errors $ r $ :
\begin{equation}
   \xi^{*}=\mathop{\arg\max}\limits_{P_{KC}} \sum r^{2}
   \label{eq5}
\end{equation}

% TODO 公式
Meanwhile, we use an ineratively re-weighted residual error function with the Huber weights function. The Huber weights are defined as:
\begin{equation}
   \delta_{\mathrm{H}}(r)=\left\{
   \begin{array}{rcl}
   1                      & & r \leq \theta_{H} \\
   \frac{\theta_{H}}{r}   & & r>\theta_{H} \\
   \end{array} \right.
   \label{eq6}
\end{equation}

% TODO 公式
The error function becomes:
\begin{equation}
   \xi^{*}=\mathop{\arg\max}\limits_{P_{KC}} \sum \delta_{H}(r) r^{2}
   \label{eq7}
\end{equation}

We use an iteratively re-weighted Levenberg-Marquardt Method\cite{ref29} to optimize Eq.(\ref{eq7}). The optimization process from coarse to fine prevents the loss function from falling into local extremes to a great extent and makes the whole VO system more robust.

\section{Resutls And Discussion}
This section tests our method on the RGB-D TUM benchmark provided by the Technical University of Munich. The RGB-D TUM benchmark is widely used by various visual mileage calculation methods. Each sequence contains RGB images, depth images, accelerometer data. There are several challenging data sets in this benchmark. For example, the duration, trajectory, translation, and rotation speed of each sequence are different. We use eight sequences to benchmark their system performance to achieve a direct comparison with ORB-SLAM2.

\subsection{Results on the TUM RGB-D Benchmark}
The methods we tested include CNN edge detection and Shi-Tomasi corner optimization (CNN+Corner), Canny edge detection and Shi-Tomasi corner optimization (Canny+Corner), dense direct method (Origin) to follow ORB-SLAM2 system comparison. In our practice, Shi-Tomasi corner optimization can significantly improve the overall accuracy of the method. When we test the data set in the Origin group, tracking loss will occur, resulting in the RPE and ATE data obtained completely deviating from the ground truth. After we added keyframes, the accuracy of the data results has also been improved. Similarly, the Canny+Corner group is also due to the addition of Shi-Tomasi corner optimization and keyframes, and the accuracy and robustness of the data results have been greatly improved. In CNN+Corner, the main method we recommend is that the test performance of data sets is excellent. The best results in multiple data sets, the sequence of estimated trajectories, and the reference trajectories of the results are shown in Figures 5, 6, and 7. On the other hand, the two groups Origin and Canny+Corner can only achieve the best results in a particular data set, and the data results in the test are not stable enough. For the ORB-SLAM2 system, because the system cannot be initialized well in some data sets, we only changed the parameter of more than 500 feature points required during initialization, and the other entire systems did not change. This means that our method just is VO to compare with the complete SLAM system.

% 表1
\begin{table*}[!htbp]
\setlength{\abovecaptionskip}{0.cm}
\setlength{\belowcaptionskip}{-0.8cm}
\caption{ RELATIVE POSE RMSE (R: DEG/S, T:M/S) OF TUM DATASETS }
\label{table_1}
\renewcommand\arraystretch{1.6}
\begin{center}
\begin{tabular}{|c|c|c|c|c|c|c|c|c|}
\hline
\multirow{2}{*}{Seq.} & \multicolumn{2}{c|}{ORB-SLAM2}                               & \multicolumn{2}{c|}{Origin}                                 & \multicolumn{2}{c|}{Canny+Corner}                           & \multicolumn{2}{c|}{CNN+Corner}                             \\ \cline{2-9} 
                      & \multicolumn{1}{c|}{RMSE(R)} & \multicolumn{1}{c|}{RMSE(T)} & \multicolumn{1}{c|}{RMSE(R)} & \multicolumn{1}{c|}{RMSE(T)} & \multicolumn{1}{c|}{RMSE(R)} & \multicolumn{1}{c|}{RMSE(T)} & \multicolumn{1}{c|}{RMSE(R)} & \multicolumn{1}{c|}{RMSE(T)} \\ \hline
fr1/xyz               & 0.978583                     & \textbf{0.015260}                     & 1.568226                     & 0.027317                     & 1.472533                     & 0.024445                     & \textbf{0.724896}                     & 0.025371                     \\ \hline
fr2/360/hemhere       & 1.145817                     & 0.127742                     & 2.872298                     & 0.390793                     & -                            & -                            & \textbf{1.021171}                     & \textbf{0.080454}                     \\ \hline
fr2/dishes            & 2.402205                     & 0.098421                     & \textbf{0.700566}                    & \textbf{0.016443}                     & 0.959770                     & 0.023250                     & 0.787372                     & 0.018816                     \\ \hline
fr2/coke              & 5.759803                     & 0.104471                     & 1.168573                     & 0.621661                     & 1.783891                     & 0.046894                     & \textbf{1.052839}                     & \textbf{0.025705}                     \\ \hline
fr3/cabinet           & -                            & -                            & 4.200325                     & 0.129285                     & 4.214291                     & 0.110508                     & \textbf{3.462629}                     & \textbf{0.081417}                     \\ \hline
fr3/large/cabinet     & 0.805515                     & 0.051905                     & 2.483856                     & 0.271248                     & 1.747718                     & 0.169891                     & \textbf{0.718709}                     & \textbf{0.048065}                     \\ \hline
fr3/str/texture/far   & 0.540715                     & 0.014733                     & \textbf{0.454445}                     & 0.013473                     & 0.781303                     & 0.019318                     & 0.455929                     & \textbf{0.013280}                     \\ \hline
fr3/str/noture/far    & 0.723704                     & 0.027528                     & 2.782048                     & 0.216976                     & 1.227557                     & 0.045992                     & \textbf{0.607414}                     & \textbf{0.021371}                     \\ \hline
\end{tabular}
\end{center}
\end{table*}

% 表二
\begin{table*}[!htbp]
\setlength{\tabcolsep}{6.8mm}
\setlength{\abovecaptionskip}{0.cm}
\setlength{\belowcaptionskip}{-0.8cm}
\caption{ ABSOLUTE TRAJECTORY RMSE(M) OF TUM DATASETS }
\label{table_2}
\renewcommand\arraystretch{1.6}
\begin{center}
\begin{tabular}{|c|c|c|c|c|}
\hline
\multirow{2}{*}{Seq.} & ORB-SLAM2  & Origin    & Canny+Corner & CNN+Corner \\ \cline{2-5} 
                      & RMSE(ATE) & RMSE(ATE) & RMSE(ATE)    & RMSE(ATE)  \\ \hline
fr1/xyz               & \textbf{0.009436}  & 0.045409  & 0.042076     & 0.038506   \\ \hline
fr2/360/hemhere       & \textbf{0.229800}  & 0.920911  & -            & 0.328613   \\ \hline
fr2/dishes            & 0.122247  & 0.094776  & 0.091492     & \textbf{0.083078}   \\ \hline
fr2/coke              & 0.419077  & 8.952361  & 0.148833     & \textbf{0.087646}   \\ \hline
fr3/cabinet           & -         & \textbf{0.389984}  & 0.447204     & 0.443365   \\ \hline
fr3/large/cabinet     & \textbf{0.090161}  & 0.346849  & 0.267330     & 0.117691   \\ \hline
fr3/str/texture/far   & \textbf{0.016497}  & 0.027270  & 0.041634     & 0.034999   \\ \hline
fr3/str/noture/far    & \textbf{0.027106}  & 0.300592  & 0.115235     & 0.036792   \\ \hline
\end{tabular}
\end{center}
\end{table*}

\subsection{Evaluation Metrics}
To measure the local accuracy of the VO method, Sturm et al.\cite{ref11}. proposed relative pose error (RPE) and absolute trajectory error (ATE). RPE measures the drift of $ \Delta t $ in a fixed time interval between a set of attitude $ Q $ from the ground true trajectory and a set of attitude $ P $ from the estimated trajectory, and the time step $ i $ is defined as:
\begin{equation}
   R P E_{i}=\left(Q_{i}^{-1} Q_{i+\Delta t}\right)^{-1}\left(P_{i}^{-1} P_{i+\Delta t}\right)
\end{equation}

where $ \Delta t $ is the time distance between poses. The ATE at a time step $ i $ is given as:
\begin{equation}
   A T E_{i}=Q_{i}^{-1} S P_{i}
\end{equation}

\begin{figure}
   \begin{center}
      \centering
      \includegraphics[width=0.42\textwidth]{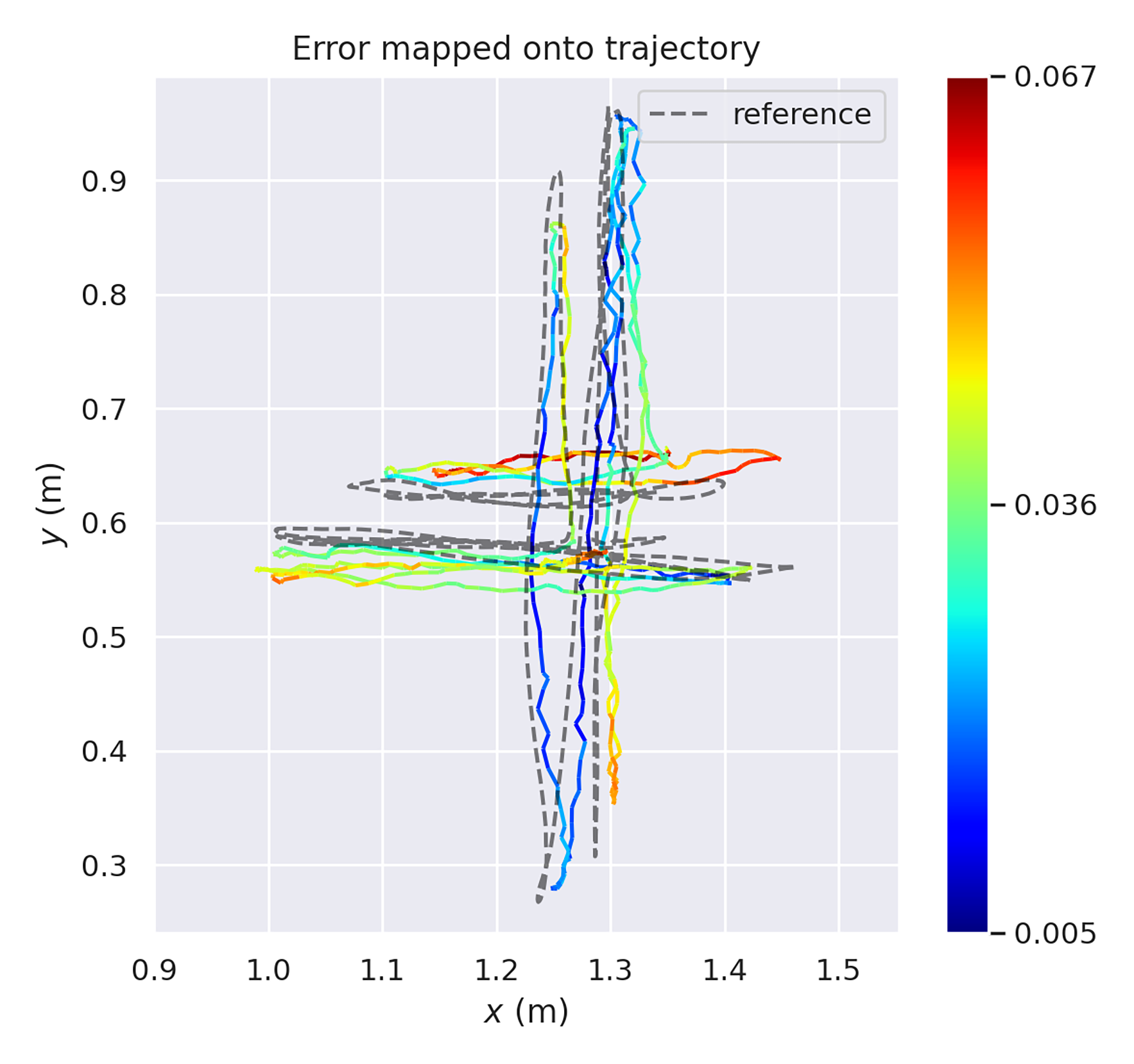}
      \caption{ The result of the sequence fr1xyz estimated trajectory and reference trajector }
      \label{fig5}
   \end{center}
\end{figure}

\begin{figure}
   \begin{center}
      \centering
      \includegraphics[width=0.42\textwidth]{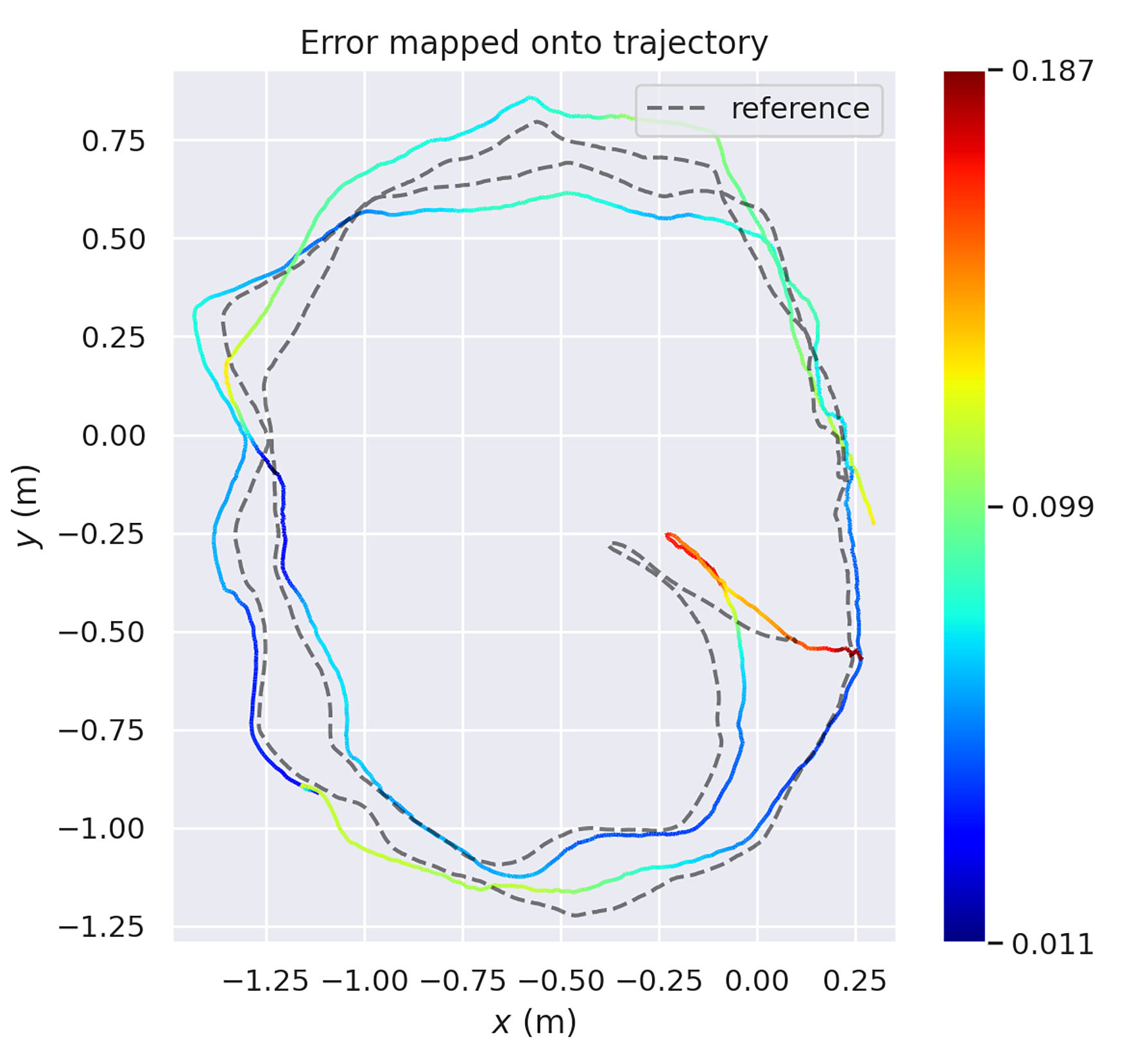}
      \caption{ The result of the sequence fr2coke estimated trajectory and reference trajector. }
      \label{fig6}
   \end{center}
\end{figure}

\begin{figure}
   \begin{center}
      \centering
      \includegraphics[width=0.42\textwidth]{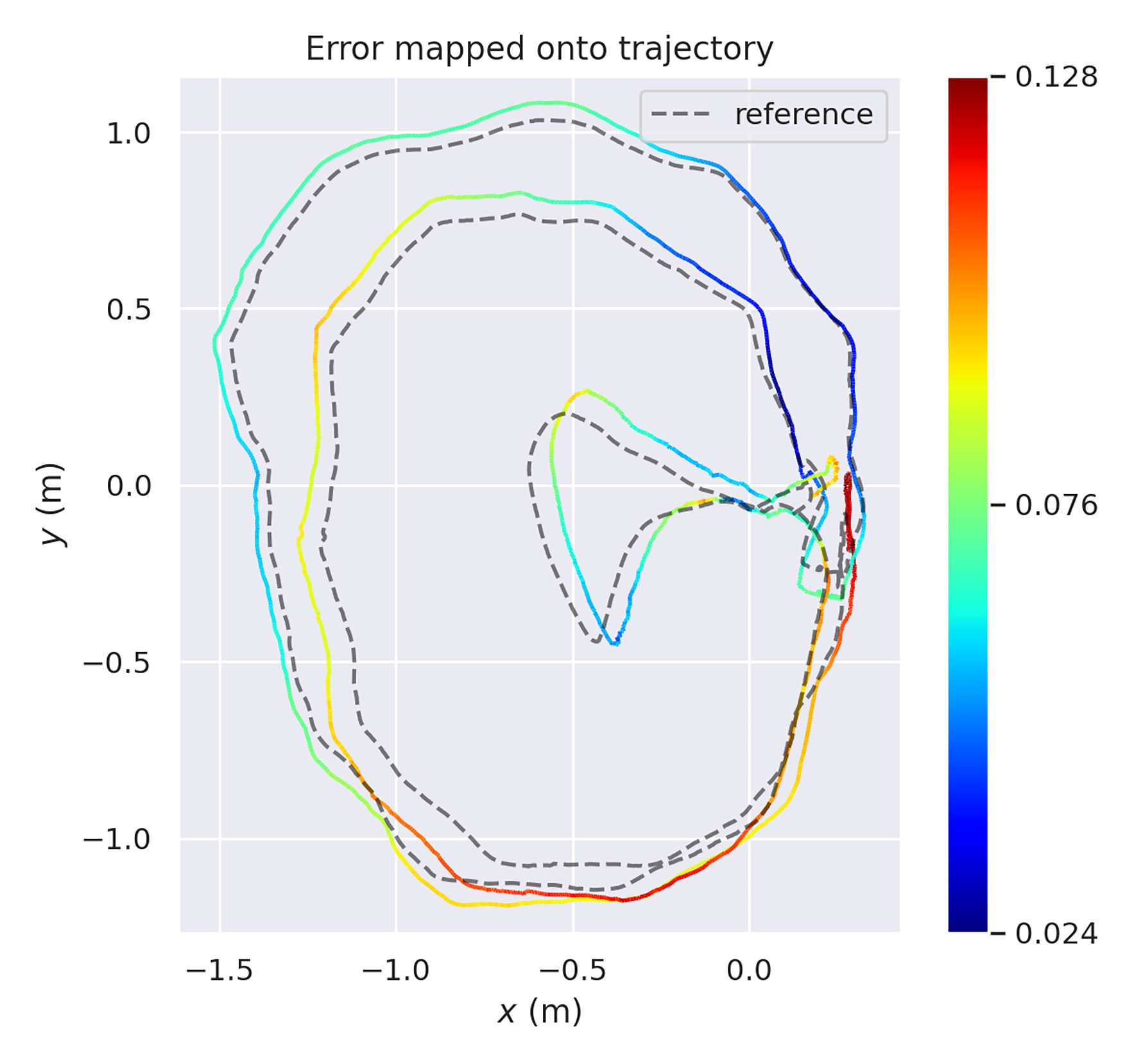}
      \caption{ The result of the sequence fr2dishes estimated trajectory and reference trajector. }
      \label{fig7}
   \end{center}
\end{figure}

where $ Q $ and $ P $ are aligned by a rigid body transformation $ S $. As suggested by Sturm et al\cite{ref31}. we evaluate the root mean squared error (RMSE) of the translational component of the RPE and ATE.

\subsection{Discussion}
From our experiments, Table \ref{table_1} can know that the RPE results of the CNN+Corner method on all data set almost entirely exceed the Canny, original image, ORB-SLAM2 system. We attribute this to the use of relatively accurate and highly robust CNN edge detection, and improve Shi Tomasi's unique corner design. At the same time, it has dual judgment mechanism key frames. From Table \ref{table_1}, we can see that the RPE data results achieve the best accuracy in multiple sequences of the data set. Even if the accuracy does not reach the best accuracy of the data results, it is only slightly inferior to the highest accuracy. This shows that our method CNN+Corner is highly accurate and robust and can work well in a variety of scenarios. From Table \ref{table_2}, we can see that the ORB-SLAM2 entire system performs very well in ATE, achieving the best accuracy in multiple data set sequences. Even though we do not use any Bundle Adjustment or global optimization as employed by ORB-SLAM2, we once again performed well in all non-SLAM methods, still showing excellent robustness and competitiveness.

\section{Conclusion}
We propose robust edge-direct visual odometry that combines CNN edge detection and Shi-Tomasi corner optimization. The LM algorithm minimizes the photometric error between the two images frames, thereby determining the relative pose of the two frames of images. Through our experiment, the method proposed in this paper is combined with direct method of Canny edge detection and ORB-SLAM2 system. Among the eight sequences, the rum rgb-d dataset has the best RPE accuracy, and the other sequences are only slightly lower. Regarding the accuracy of ATE, our method also has an excellent performance in non-slam methods. Although compared with the entire system of SLAM, our method can still achieve the best in two sequences without back-end optimization. Our method can run well in multiple scenarios, demonstrating the accuracy and robustness of our method. There is still a gap between the accuracy of our proposed method in ATE and the entire ORB-SLAM2 system. We will add back-end optimization to our method in the future to perfect the whole SLAM system.

%
%You may put all reference items in a separate file, say myRef.bib, in bibTex format.
%
%\bibliographystyle{IEEETran}
%\bibliography{myRef} %change to your file name (with suffix .bib)

\end{document}